\documentclass[12pt]{iopart}
\usepackage[utf8]{inputenc}
\usepackage{float}
\usepackage{setstack}
\usepackage{siunitx}
\usepackage{graphicx}
\usepackage{caption}
\usepackage{subcaption}
\usepackage{hyperref}
\usepackage{numprint}
\usepackage{mfirstuc}
\usepackage{microtype}
\usepackage{parskip}
\usepackage{iopams}
\usepackage{xcolor}
\MFUnocap{are}
\MFUnocap{or}
\MFUnocap{etc}
\MFUnocap{of}
\MFUnocap{in}
\MFUnocap{and}
\providecommand{\keywords}[1]{\noindent{\it{Keywords}}: #1}

\begin{document}
\title[Learning to See via Epiretinal Implant Stimulation \textit{in silico} with Model-Based Deep Reinforcement Learning ]{Learning to See via Epiretinal Implant  Stimulation \textit{in silico} with Model-Based Deep Reinforcement Learning }

\author{Jacob Lavoie$^1$, Marwan Besrour $^1$, William Lemaire$^1$, Jean Rouat$^1$, Réjean Fontaine$^1$, Eric Plourde$^1$}

\address{$^1$ Department of Electrical Engineering and Computer Engineering, Université de Sherbrooke, Sherbrooke, Quebec, J1K 2R1, Canada}

\begin{abstract}

\noindent Objective: Diseases such as age-related macular degeneration and retinitis pigmentosa cause the degradation of the photoreceptor layer.
One approach to restore vision is to electrically stimulate the surviving retinal ganglion cells with a microelectrode array such as epiretinal implants.
Epiretinal implants are known to generate visible anisotropic shapes elongated along the axon fascicles of neighboring retinal ganglion cells.
Recent work has demonstrated that to obtain isotropic pixel-like shapes, it is possible to map axon fascicles and avoid stimulating them by inactivating electrodes or lowering stimulation current levels.
Avoiding axon fascicule stimulation aims to remove brushstroke-like shapes in favor of a more reduced set of pixel-like shapes. 
Approach: In this study, we propose the use of isotropic and anisotropic shapes to render intelligible images on the retina of a virtual patient in a reinforcement learning environment named \href{https://github.com/NECOTIS/rlretina.git}{rlretina}.
The environment formalizes the task as using brushstrokes in a stroke-based rendering task.
Main Results: We train a deep reinforcement learning agent that learns to assemble isotropic and anisotropic shapes to form an image.
We investigate which error-based or perception-based metrics is adequate to reward the agent.
The agent is trained in a model-based data generation fashion using the psychophysically validated axon map model to render images as perceived by different virtual patients.
We show that the agent can generate more intelligible images compared to the naive method in different virtual patients.
Significance: This work shares a new way to address epiretinal stimulation that constitutes a first step towards improving visual acuity in artificially-restored vision using anisotropic phosphenes.

\end{abstract}

\keywords{deep reinforcement learning, model-based, retinal implant, stroke-based rendering}

\maketitle

\noindent{\small This is the authors' preprint version. The published article is: J. Lavoie, M. Besrour, W. Lemaire, J. Rouat, R. Fontaine and E. Plourde, ``Learning to See via Epiretinal Implant Stimulation \textit{in silico} with Model-Based Deep Reinforcement Learning,'' \textit{Biomedical Physics \& Engineering Express} \textbf{10}, 025006 (2024). DOI: \href{https://doi.org/10.1088/2057-1976/acf1a5}{10.1088/2057-1976/acf1a5}.}

\ioptwocol

\section{Introduction}

Vision loss has a serious impact on quality of life.
Epidemiological studies reveal that vision loss also has an important global burden of disease on society \cite{GDB_children_lancet}.
There are effective treatments for common eye diseases such as myopia, glaucoma, and cataracts.
However, treatments aimed at prevalent diseases affecting the retina, such as age-related macular degeneration and retinitis pigmentosa, can slow the disease at best.
Age-related macular degeneration accounts for 15.85\% of incurable vision loss cases \cite{GDB_children_lancet}.
Therefore, it is the most prevalent untreatable disease that causes vision loss in developed countries \cite{GDB_children_lancet}.
Retinitis pigmentosa is a rare disease with a worldwide prevalence of 1/4000 that causes vision impairment to complete loss during adolescence and young adult life \cite{retinite_lancet}.
The early onset of retinitis pigmentosa increases the detrimental burden of the disease and remains one of the leading causes of blindness in the 20-year-old to 64-year-old age group \cite{scandinavian_retinite_ophatamology,10.1001/archopht.116.5.653}.

Age-related macular degeneration and retinitis pigmentosa cause degeneration of the retina's photoreceptor layer.
Therefore, patients gradually lose their sensitivity to light, leaving subsequent layers of neurons with an aberrant signal.
One treatment is to electrically stimulate the surviving neurons to artificially restore a certain visual acuity.
This can be performed with the use of microelectrode arrays (MEA) that can be implanted to target different layers of the retina \cite{chuang2014retinal}.
These devices were developed based on the observation that focal electrical stimulation of the retina generates a dot-shaped visual perception called phosphene \cite{humayun1996visual}.
Phosphenes are spatially preserved along the visual pathway as a result of the retinotopic organization of the visual system.
An ensemble of phosphenes caused by retinal stimulation is referred to as a percept.
The main focus of the work presented in this article is to train a reinforcement learning (RL) agent that selects phosphenes to generate a percept similar to a digital image in different virtual patients, thus restoring visual acuity.
Before presenting the proposed approach, we go through the train of thought that led to our attempt to leverage anisotropic phosphenes instead of mitigating them.

\subsection{Stimulation sites}
\label{ssec:recent_stimulation_approaches}
\begin{figure}[t]
    \centering
    \includegraphics[width=0.5\textwidth]{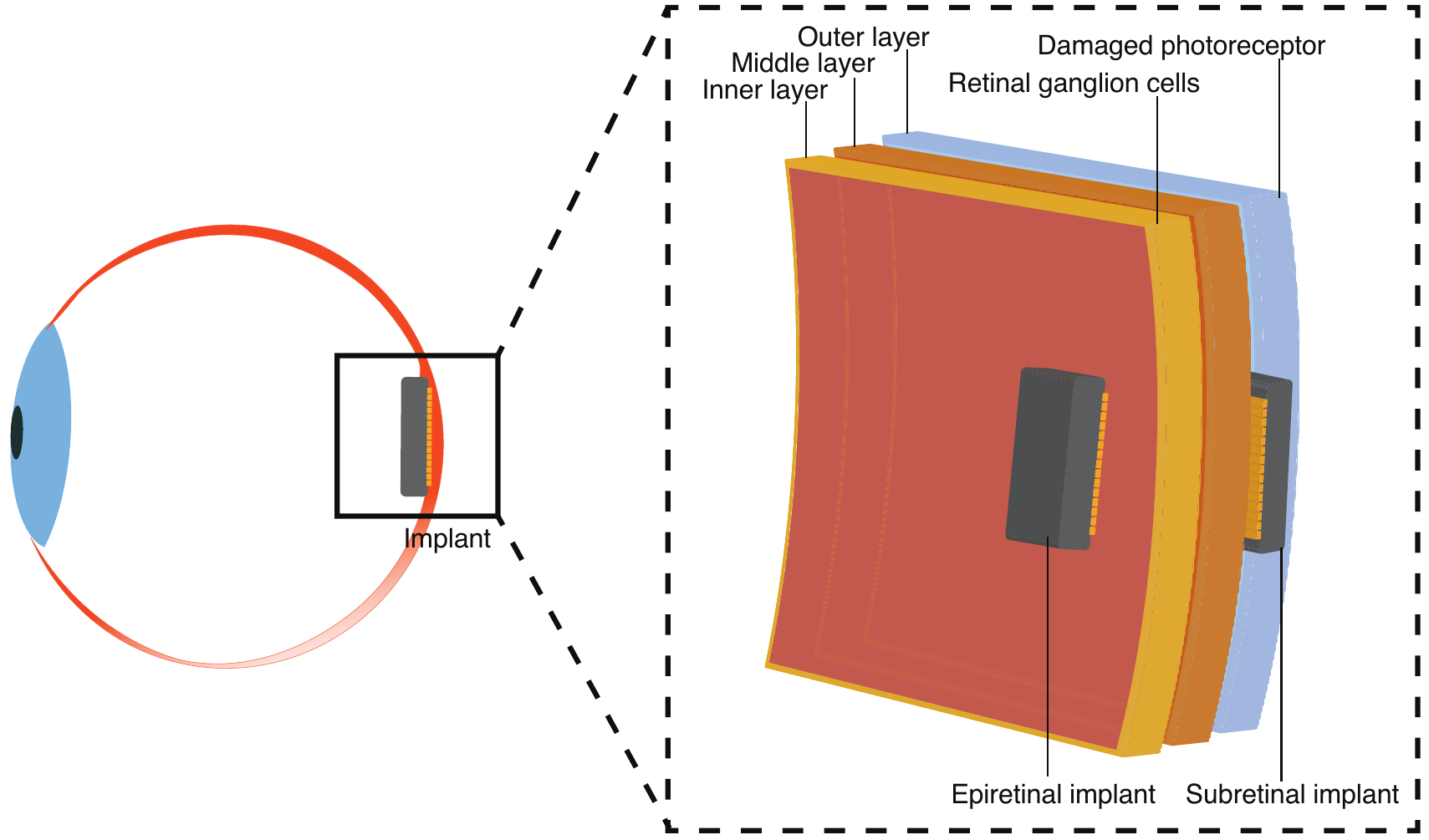} %
    \caption{Epiretinal and subretinal implantation sites are illustrated in the anatomical context of the retina.}
    \label{fig:implant_site}
\end{figure}

There are two MEA implantation sites, shown in \fref{fig:implant_site}, which are often found in the literature \cite{chuang2014retinal,zrenner2002will}: the subretinal and epiretinal implantation sites.
In addition to the aforementioned sites, suprachoroidal, lateral geniculate nucleus, and visual cortex implants are also potential stimulation sites that target different parts of the visual pathway \cite{kleinlogel2020emerging}.
The epiretinal implantation site stimulates retinal ganglion cells (RGC), while the subretinal implantation site stimulates bipolar cells.
Subretinal implants hold great promise in terms of visual acuity as a result of the lower-level signal encoded by bipolar neurons compared to subsequent layers \cite{palanker2022simultaneous}.
However, the insertion point makes its deployment more complex.
In fact, to be installed, the subretinal implant must be surgically implanted between the pigment epithelium and the outer retina.
Therefore, its use is limited to patients with intact inner and middle layers of the retina \cite{pavlova2019epiretinal}.
In addition, data and power are generally transmitted through wires and an induction coil to the subretinal implant.
Epiretinal implants can be placed on the retinal surface, allowing the use of optical power and data transmission \cite{lemaire2021retinal,zrenner2002will}.
In this study, we focus mainly on epiretinal implant stimulation due to its potential to help more patients and be less invasive.

\subsection{Psychophysical study of anisotropic phosphenes in epiretinal stimulation}
\label{ssec:phenomenological}
\begin{figure}[t]
    \centering
    \includegraphics[width=0.5\textwidth]{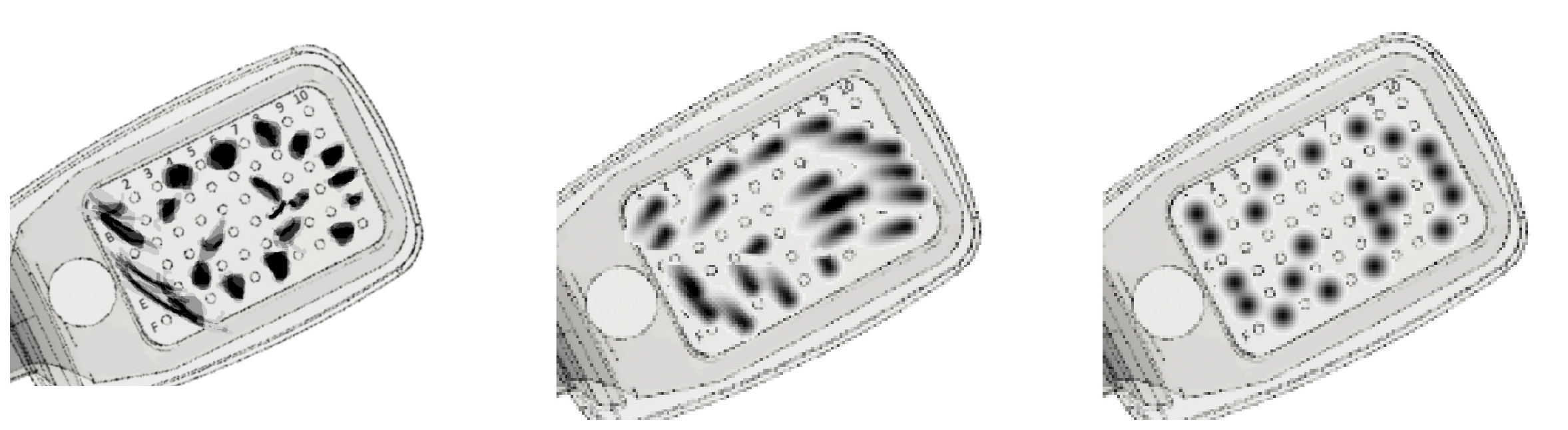}
    \caption{Comparison between anisotropic and isotropic phosphenes shapes. (Left) Subject 2 drawings from \cite{beyeler2019model} show anisotropic phosphenes perceived during single-electrode stimulation with an Argus II implant. (Center) Axon map model predicts the anisotropic shaped caused by extracellular axon stimulation. (Right) Scoreboard model does not include extracellular axon stimulation resulting in ideal isotropic phosphenes. Phosphenes and images are generated using the \textit{pulse2percept} Python module \cite{beyeler2017pulse2percept}.}
    \label{fig:anisotropic_phosphene}
\end{figure}

Recent experiments with patients using an epiretinal implant revealed that phosphenes are not always dot-shaped \cite{beyeler2019model}.
In fact, although the epiretinal implant aims to stimulate the RGCs, it also stimulates the peripheral axons of the RGCs that are sufficiently close to the stimulating electrode.
The stimulation of peripheral RGC axons causes the patient's visual system to render irregular shapes as shown in \fref{fig:anisotropic_phosphene} \cite{tsai2012responses, beyeler2019model}.
Axons are organized in fascicles, also known as axon bundles, which converge to the blind spot to form the optic nerve.
Epiretinal electrical stimulation, therefore, generates an elongated shape parallel to the axon fascicles \cite{beyeler2019model} as shown in \fref{fig:anisotropic_phosphene}.
The phosphene is elongated along the axons, causing a perfect isotropic phosphene to become anisotropic.

\subsection{Naive stimulation algorithm}
\label{ssec:nsa_stimulation_algorithm}

The Naive Stimulation Algorithm (NSA) for the epiretinal implant described in \cite{luo2016argus} is tested in patients with the Argus II epiretinal implant. 
Many improvements to the original algorithms that use more advanced image processing methods such as Difference of Gaussian \cite{guo2023edge}, constrained optimization \cite{spencer2019global}, and deep learning \cite{wu2023deep,granley2022hybrid} obtained better results in simulation \cite{borda2022advances}.
The NSA algorithm for epiretinal implant most tested in patients is transforming a digital image into electrical stimulation using a downscaling operation of a camera image that matches the intensity of a pixel to the amplitude or frequency.
This results in a pixelated image having the dimension of the microelectrode array.
Electrical stimulations are proportional to the intensity of the pixels. 
These electrical stimulations are delivered in a temporal sequence of single-electrode stimulations in an experimental setup preventing eye movement.
Simultaneous stimulation with multiple electrodes causes more irregular phosphene shapes \cite{rizzo2003perceptual}.
Therefore, the NSA stimulation algorithm does not consider the fact that RGC axons are stimulated, leading to anisotropic phosphene \cite{luo2016argus}.

\subsection{Mitigating axon bundle stimulation}\
\label{ssec:mitigation}
As indicated above, axon bundle stimulation produces anisotropic phosphenes.
Previous work attempted to minimize the impact of anisotropic shapes on the percept quality.
One such approach consists of modifying the electrode configuration to attenuate this impact \cite{esler2018minimizing,bruce2022greedy}.
Other approaches adopt different stimulation techniques, such as RGC mapping and current steering, creating virtual electrodes between electrodes, resulting in more consistent isotropic phosphenes with a healthy retina \textit{in situ} \cite{jepson2014spatially,grosberg2017activation,tandon2021automatic,vilkhu2021spatially,tong2019improved,Gonzalez2016,chang2019stimulation,ghaffari2020effect}.
These approaches offer better control over RGC spiking and collateral axon stimulation.
In addition, both tend to limit the number of usable electrodes and the range of possible stimulation intensity, thus reducing the diversity of shapes that can be generated by epiretinal implants \cite{grosberg2017activation,bruce2022greedy}.

\subsection{Leveraging axons bundle stimulation with stroke-based rendering}

Data acquired from patients with epiretinal implants and anatomical studies allow the development of a psychophysically validated model of end-to-end visual processing in the degenerated retina \cite{beyeler2019model}.
These models help to visualize the perceived anisotropic shapes created by stimulation of axon bundles in an epiretinal stimulation setting, as shown in \fref{fig:anisotropic_phosphene}.

Instead of designing a retinal stimulation algorithm that mitigates anisotropic shapes, it is possible to use the available shapes produced by all possible stimulations to form the desired image to be perceived.
Choosing anisotropic shapes to generate an image is referred to in the computer vision community as stroke-based rendering (SBR).
More precisely, SBR is a non-photorealistic method to create imagery from discrete elements called strokes, such as paint strokes or ripples \cite{hertzmann2003survey}.
An analogy that illustrates the problem is that of the painter reproducing a photograph on a canvas.
In this paper, the phosphene is considered equivalent to one brushstroke, and the retina is the canvas.

\subsection{Reinforcement learning in stroke-based rendering}
Recent successful SBR algorithms such as SPIRAL \cite{spiral}, StrokeNet \cite{zheng2018strokenet}, Doodle-SDQ \cite{Zhou2018LearningTD}, Sketch-RNN \cite{sketchrnn} and other model-based approaches \cite{huang2019learning} showed high-quality picture reproduction. 
All successful attempts mentioned the use of deep reinforcement learning (DRL) paradigm for paint stroke decomposition and/or Generative Adversarial Networks (GAN) for quality assessment.
Thinking of epiretinal stimulation as an SBR problem allows us to train a deep reinforcement learning agent to use the full diversity of shapes produced by all electrodes.
This new perspective offers the use of anisotropic shapes rather than mitigating them as in the aforementioned approaches in \sref{ssec:mitigation}.

\subsection{Reinforcement learning and retinal stimulation}
\label{ssec:reinforcement_learning_retinal_stimulation}
Previous work using reinforcement learning (RL) to adjust epiretinal stimulation parameters used the center-surround RGC receptive field model as a premise to simulate the retina \cite{becker1998retina,eckmiller2005tunable}.
The center-surround RGC receptive field is the accepted model for RGC neural coding.
It consists of a circular area of the retina called the center and the surrounding region that respond oppositely to light exposure.
However, it does not consider the stimulation of axon bundles in the context of epiretinal stimulation.
Some more recent work using DRL is promising but does not include anisotropic phosphene in the percept generation \cite{kuccukouglu2022optimization}.
Nonetheless, as the author notes in \cite{becker1998retina}, RL is a very appealing framework to use patient's feedback as a learning signal to automate adaptation to different patients \textit{in vivo} \cite{becker1997spatio}.

For each patient, the best ensemble of phosphenes that render a particular target image on the retina is not known.
The current virtual patient models mentioned in \fref{fig:anisotropic_phosphene} can only generate a percept from electrode stimulation.
One way to find the best electrode stimulation combination from a target image is to evaluate the perception of the predicted percept compared to the target image.
A brute-force approach could be to try every combination of electrodes, generate the percept, and then calculate the similarity between the percept and the target image.
This approach becomes more tedious as the state-space, or in this case, the number of electrodes increases in complexity \cite{barto2004reinforcement}.
Therefore, it is difficult to generate a complete dataset to leverage supervised learning methods.
RL allows for searching the state-space of electrode combinations more efficiently \cite{barto2004reinforcement}.
A DRL agent that interacts with an environment gets direct feedback through a reward as it approaches a solution.
The agent learns to associate a particular state of the environment with an action it can take.
RL is therefore a suitable paradigm for epiretinal stimulation because there exists a model of epiretinal stimulation with which an agent can interact, but there is no optimal solution to transform a target image into an electrode combination.

\subsection{Proposed approach}
This work bridges the gap between the previous attempt \cite{becker1998retina} to find optimal stimulation parameters with an epiretinal implant using RL and the latest anatomical knowledge and lessons learned from trials of human epiretinal implants \cite{beyeler2019model}.
It is the only attempt to improve visual acuity for the implanted patient by leveraging anisotropic phosphene in a SBR problem.

More specifically, we want to investigate whether a DRL agent can learn to generate a sequence of single-electrode stimulation from a target image, thus increasing visual acuity for different virtual patients with an epiretinal implant. This paper proposes the following contributions:

\begin{itemize}
    \item We formalize the transformation of the original image into a stimulation pattern as an SBR problem implemented in a new reinforcement learning environment named \textit{rlretina} available at \url{https://github.com/NECOTIS/rlretina.git}.
    \item We compare a pixel-based distance and distance between probability distributions as a proxy of a perceptual metric in the reward design of the new environment.
    \item We build a model-based DRL agent that can explore the available anisotropic phosphene space to increase visual acuity in virtual patients with different epiretinal implant settings.
    \item Finally, we investigate perceptual metrics to circumvent pixel-based metric limitations to better compare the percepts resulting from different algorithms.
\end{itemize}

In addition, we demonstrate pixel-based error limitations as a reward in this perceptual task.
We compare the proposed DRL agent with the NSA stimulation algorithm using the mean structural similarity index measure (MSSIM) as an evaluation metric presented in \sref{sssec:experiment_different_patient}.
We show that the proposed agent better reproduces the original images in the virtual patient's percepts than the NSA stimulation methods.
It also performs better with different virtual patient implant settings than the NSA stimulation algorithm.

\subsubsection{Using a model-based approach}
\label{sssec:model_based_approach}

\begin{figure}[t]
    \centering
    \includegraphics[width=0.5\textwidth]{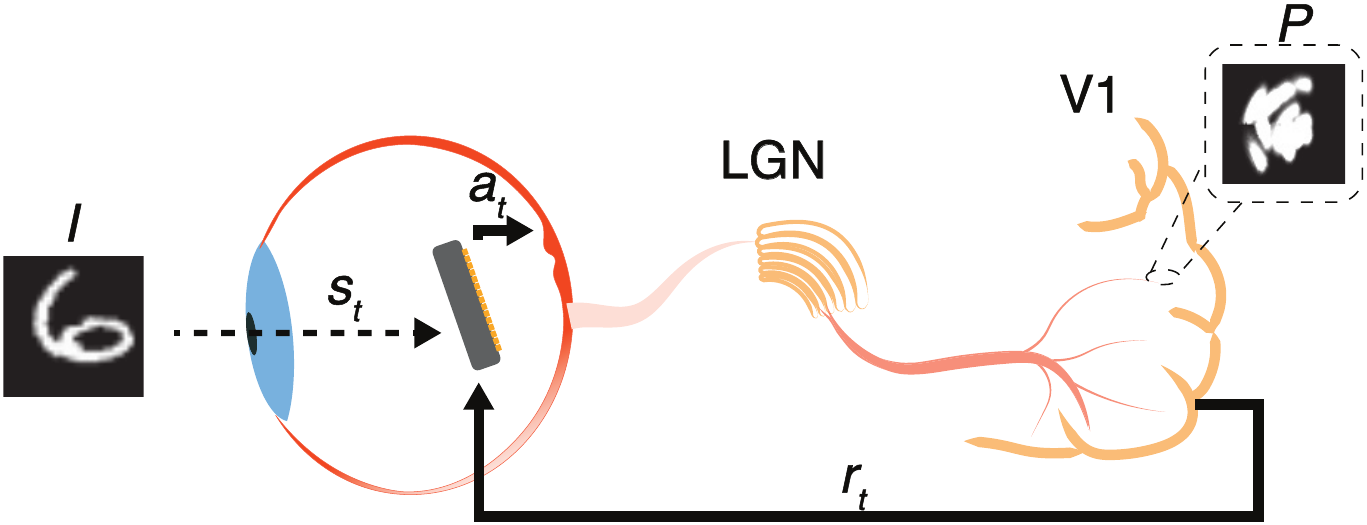}
    \caption{A model-free environment simulates a virtual patient's retina and its visual system. The agent is defined as the implant controller and is in charge of transforming a camera image into a single-electrode stimulation pattern. The single-electrode stimulation excite the retina. The signal propagates to the lateral geniculate nucleus (LGN) and through the visual cortex (V1) and subsequent visual areas (Not illustrated). The environment returns a state $s_t$. The agent must interact with the environment through an action $a_t$ corresponding here to the electrode's stimulation. The environment then gives a reward $r_t$ according to the similarity between the target image $I$ and the visual perception $P$ resulting from the electrical stimulation.}
    \label{fig:implant}
\end{figure}

A model-free RL approach to epiretinal stimulation in an \textit{in vivo} system is presented in \fref{fig:implant}.
The state $s_t$ contains only the digital image $I$.
It is the reward $r_t$ that indirectly provides information about the similarity between $I$ and the percept $P$. 
The virtual patient's visual system illustrated in \fref{fig:implant} is conceptualized as part of the environment (see \sref{sssec:environment}).
This approach was used in previous work presented in \sref{ssec:reinforcement_learning_retinal_stimulation}.

In the present work, we adopt a model-based approach to data generation explained in \sref{sssec:agent} by including $P$ in $s_t$ as shown later in \fref{fig:model_based_agent}.
Using an external world model to generate the data for the agent, in this case, the percept $P$, is derived from the recurrent World Models proposed in \cite{recurrentWorldModels}.
It alleviates the agent's burden of modeling the environment that can be composed of any real-world image.
Instead, it uses the axon map model to generate the image.

\subsubsection{Epiretinal stimulation as Markov Decision Process}
\label{sssec:stimulation_as_mdp}
A Markov Decision Process (MDP) allows one to formalize the interaction between the agent and the environment in a RL task.
The proposed approach is inspired by the way the SBR task is formalized as a MDP where the goal is to maximize the similarity between a target image and a stroke-based rendered image equivalent respectively to $I$ and $P$ in \fref{fig:implant}.
Selecting single-electrode stimulation in a sequence is a decision-making task that can be modeled as a MDP.
A finite MDP is a decision-making process that satisfies the Markov property stipulating that the action influences not only the immediate reward $r_t$, but also the probability that the process moves into its new state $s'$ at $t+1$ \cite{sutton_barto_2018}.
A finite MDP is defined by sets with a finite number of elements for states $\mathcal{S}$, actions $\mathcal{A}$, and rewards $\mathcal{R}$.
Given $s_t$ the state at time $t$, and $a_t$ the action at time $t$, the dynamic or state transition function of the environment $p$ is defined in \eref{eq:mdp} \cite{sutton2018reinforcement}.

\begin{equation}
    \label{eq:mdp}
    p(s', r|s,a) \doteq Pr\{s_{t+1} = s', r_{t+1} = r|s_{t},a_{t}\}
\end{equation}

for all $s'$ where $s \in \mathcal{S}$, $r \in \mathcal{R}$ and $a \in \mathcal{A}(t)$.
In other words, the next state $s_{t+1}$ and the associated reward $r_{t+1}$ are functions of the probability $Pr$ of taking action $a_t$ in the previous state $s_t$.
The state $s_t$ given to the agent is defined as the target image $I$, and the percept $P$ that exposes the internal dynamics of the retina model.
The agent must then decide on an action $a_t$, namely, a single-electrode stimulation.
As mentioned previously, single-electrode stimulation generates anisotropic shapes perceived here by the virtual patient.
The percepts of the virtual patient are simulated with a psychophysically validated model developed from a human patient implanted with an Argus epiretinal implant \cite{beyeler2019model}.
This model of human epiretinal vision is used by the environment that simulates a virtual patient.

\section{Materials and methods}
\label{sec:materials_methods}
The following sections present the implant stimulation algorithm and the experiences that result in the generation of intelligible percepts.
The algorithm takes the form of a DRL agent detailed in \sref{sssec:agent} estimating the best sequence of action $a_t$ to maximize the similarity between $I$ and $P$ to increase the visual acuity of the patient.

\subsection{Materials}
\label{ssec:materials}

\subsubsection{Environment}
\label{sssec:environment}

We developed a new environment to train reinforcement learning agents.
The environment follows the OpenAI gym specifications \cite{openai2016git}.
\Sref{sssec:environment} presents the underlying assumptions in the environment.

\paragraph{From single-electrode stimulation to percept}

The environment uses the axon map model that accurately reproduces the drawings of the phosphene perceived by real patients during an epiretinal single-electrode stimulation \cite{beyeler2019model}.
It is important to note that the axon map model was developed in a control clinical set-up that prevented eye movements and used only single-electrode stimulation without superimposition of multiple stimulations.
Therefore, it serves as an anisotropic phosphene rendering tool to simulate the virtual patient.
The axon map model, detailed in the section below, renders anisotropic phosphenes as observed in implanted patients rather than ideal isotropic phosphenes.
Rendered phosphenes are assembled to form the percept $P$.
In the case at hand, an episode of agent-environment interaction is defined as a sequence of single-electrode stimulations.
At the beginning of the episode, a new target image $I$ is selected from the image dataset.
At each step or single-electrode stimulation of an episode, the agent produces a vector with probabilities of selecting each electrode.
The number of activated electrodes and their normalized stimulation values are set in the environment configuration.
The environment uses single-electrode stimulation equivalent to the action $a$ of the agent to update the virtual patient's percept $P$.

\paragraph{Axon map model}

An electrical stimulation is assumed to generate focal dots of light that decay exponentially with the distance between the location of the stimulating electrode and the location of the stimulated retina $(x_{stim}, y_{stim})$ and the spatial decay constant $\rho$.
These assumptions are included in the calculation of the scoreboard model to estimate the intensity profile $I_{score}(x, y; \rho)$ \cite{beyeler2019model}.
Equation \eref{eq:score} corresponds to the scoreboard model presented in \fref{fig:anisotropic_phosphene} \cite{beyeler2019model}.

\begin{equation}
\resizebox{0.45\textwidth}{!}{$
    I_{score}(x, y; \rho) = \exp \Big(-\frac{(x-x_{stim})^2 + (y-y_{stim})^2}{2 \rho^2} \Big)
    \label{eq:score}
$}
\end{equation}

The axon map model estimates the contribution of axon stimulation to the virtual patient's perception based on anatomic observation of axonal growth \cite{JANSONIUS20092157}.
The axonal trajectories are represented with a modified polar coordinate system with its center at the optical disc.
The contribution of axon stimulation to a phosphene decays exponentially along the axon bundles with the distance between the location of the stimulating electrode and the soma $(x_{soma}, y_{soma})$.
The impact of axon stimulation on the intensity profile $I_{axon}(x,y; \lambda)$ is estimated using \eref{eq:axon}. 

\begin{equation}
\resizebox{0.45\textwidth}{!}{$
    I_{axon}(x,y; \lambda) = \exp \Big(-\frac{(x-x_{soma})^2 + (y-y_{soma})^2}{2 \lambda^2} \Big)
    $}\
    \label{eq:axon}
\end{equation}

where $\lambda$ is a constant that modulates spatial decay along the axon.
Therefore, we can combine \eref{eq:score} and \eref{eq:axon} to predict the intensity profile $I_{map}(x,y; \rho, \lambda)$ of anisotropic phosphenes perceived by virtual patients implanted with an epiretinal implant:

\begin{eqnarray}
    I_{map}&(x,y; \rho, \lambda) = I_{score}(x, y; \rho)I_{axon}(x,y; \lambda)
\end{eqnarray}

The values of parameters $\rho$ and $\lambda$ can be set to simulate implant placement relative to the retina of different virtual patients, as demonstrated with real patients in \cite{beyeler2019model}.

\paragraph{Implant simulation}

A model of the commercialized ArgusII implant \cite{luo2016argus} is used in the simulations presented in the current work to facilitate the reproducibility of the experiments and further comparisons with other stimulation algorithms.
The implant is placed on the surface of the retina according to the coordinates centered on the fovea.
The angle of insertion is set through the environment configuration.
The implant placement parameters of subjects in \cite{beyeler2019model} are also available in the environment.
All single-electrode pulse waveforms consist of a biphasic, cathodic-first, charge-balanced, square-wave pulse.

\paragraph{Reward definition}

The reward is defined as follows:

\begin{equation}
    r(s_t, a_t) = \frac{L_{t+1} - L_t}{L_{t_0}}
    \label{eq:reward}
\end{equation}

where $L_t$ is a given distance between the target image $I$ and the percept $P$ at time $t$ and $t_0$ indicates the beginning of the experiment.
The difference in two subsequent time steps, $L_{t+1} - L_t$, is used to signal the agent if it is getting closer or farther from the target.

\paragraph{Dataset}

The MNIST \cite{deng2012mnist} dataset used in the experiments serves as a proxy for visual acuity tasks often used in optometry, such as the Snellen chart \cite{snellen1873probebuchstaben}.
It contains 70 000 28 $\times$ 28 handwritten number images split into a 60 000 image training set and a 10 000 image test set.

\subsubsection{Agent}
\label{sssec:agent}

\paragraph{Building of a model-based agent}
\hfill

The agent architecture is the Soft Actor-Critic (SAC) algorithm \cite{haarnoja2018soft}.
An actor-critic agent is made up of two parts; (1) the actor who learns a policy $\pi(a_{t}|s_{t})$ that maps a state $s_t \in \mathcal{S}$ to actions $a$ and (2) the critic who approximates a value function that gives an evaluative feedback based on the agent's action $a_t$ in state $s_t$.
The SAC algorithm uses a Q-function in the critic in a similar way to recent agent algorithms such as the Deep Deterministic Policy Gradient (DDPG)\cite{silver2014deterministic,lillicrap2019continuous}.
This allows the agent to be trained in an off-policy manner and, therefore, reuse data efficiently compared to the standard policy iteration used in the classic actor-critic formulation \cite{sutton_barto_2018}.
The SAC algorithm uses a stochastic actor and maximizes the entropy of the actor with an entropy maximization objective.
This results in a more stable and scalable algorithm that exceeds the efficiency and final performance of DDPG \cite{haarnoja2018soft}.
The associated SAC cost function is as follows :

\begin{equation}
    J(\pi) = \sum_{t=0}^{T}\mathbb{E}(s_t,a_t)_{\sim \rho_\pi}[r(s_{t} , a_{t}) + \alpha H(\pi (a_t|s_{t}))]
    \label{eq:sac_loss}
\end{equation}

where the parameter $\alpha$ controls the stochasticity of the optimal policy during training and $\mathbb{E}$ denotes the mathematical expectation \cite{haarnoja2018soft}. 
Like in standard RL, the SAC algorithm maximizes the expected sum of rewards $\sum_{t=0}^{T}\mathbb{E}(s_t,a_t)_{\sim \rho_\pi}[r(s_{t} , a_{t})]$.
As mentioned above, the SAC algorithm also includes the expected entropy $\sum_{t=0}^{T}\mathbb{E}(s_t,a_t)_{\sim \rho_\pi}[H(\pi(\dot|s_{t}))]$ of the policy over $\rho_{\pi}(s_{t})$ in the loss function $J(\pi)$ as shown in \eref{eq:sac_loss}.
In the proposed approach, both the actor and the critic are approximated using a convolutional neural network (CNN) with a CoordConv \cite{liu2018intriguing} layer (See \Tref{tab:hyperparameters}).

\begin{figure}[t]
    \centering
    \includegraphics[width=0.5\textwidth]{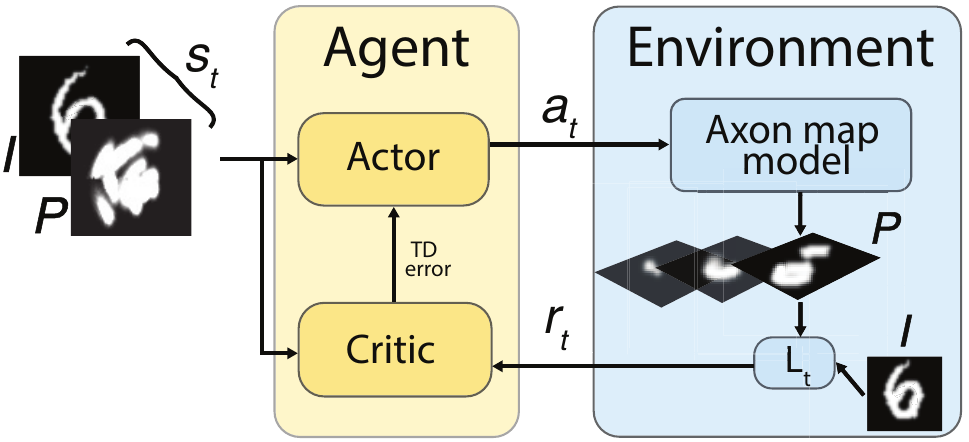}
    \caption{In contrast to \fref{fig:implant}, the SAC agent receives the percept $P$ as well as the original image $I$. This model-based data generation scheme makes use of the predictive model of the retina to give access to the internal state $L_t$ of the environment model. Electrical stimulations are transformed into a visual percept using the axon map model from \textit{pulse2percept} library. The SAC agent is composed of an actor and a critic that takes $I$ and $P$ as input $s_t$. The actor is in charge of estimating the action $a_t$ equivalent to a single-electrode stimulation. The critic gives an evaluative feedback to the actor taking action $a_t$ at state $s_t$. }
    \label{fig:model_based_agent}
\end{figure}

As illustrated in \fref{fig:model_based_agent}, the actor and the critic take the state $s_t$ composed of the target image $I$ and the percept $P$ at time $t$.
The agent does not need to model the environment implicitly, as opposed to a model-free approach.
The transition dynamic of the environment corresponds to the axon map model that generates the percept $P$.
The percept is then given directly to the agent in $s_t$.
Therefore, the agent uses model-based data generation.

\paragraph{Training of the agent}

This section presents the details of the agent's training implemented with the \textit{Ray RLlib} deep reinforcement learning library \cite{liang2018rllib}.
All hyperparameters to replicate the agent are collected in \tref{tab:hyperparameters}.
The agent interacts with the environment until it reaches the $N$ steps corresponding to $N$ single-electrode stimulation.
$N$ steps form an episode.
The agent is trained \textit{tabula rasa} with batches of 32 episodes over 1000 iterations.
The agent's replay memory buffer that allows for more stable learning and off-policy training is set to hold the latest $10^6$ steps \cite{mnih2015human}.
The Adam \cite{kingma2014adam} optimizer is used to train the neural networks of the actor and the critic.
The agent is trained on two AMD Milan 7413 with 24 cores running at 2.65 GHz and one NVidia A100 GPU.

\Table{\label{tab:hyperparameters}SAC hyperparameters.}
         \br
         Parameters & Values \\
         \mr
         Learning rate & $3 \cdot 10^{-4}$ \\
         Discount factor ($\gamma$) & 0.99 \\
         Replay buffer & $10^{6}$ \\
         Number of hiddens layers & 3 \\
         Number of hidden units by layers & 512 \\
         Number of samples by minibatch & 32 \\
         Nonlinearity & ReLU \\
         Reward scale & 200 \\
         Target smooth coefficient ($\tau$) & $0.0005$ \\
         Target update interval & $1$ \\
         \br
\endTable  

\subsection{Methods}
\label{ssec:methods}

To validate the reward design of the environment and compare the agent with the NSA approach, two experiments were carried out (1) a pivotal experiment using pixel-based and perception-based metrics as reward to obtain a readable percept produced by the agent and (2) an experiment that demonstrates the agent's ability to adapt to different patients.

\subsubsection{Effect of reward shaping on agent’s learning}
\label{sssec:experiment_reward_shaping}

Having a suitable metric to measure the pixel and perceptual similarities between the percept and the target image is critical to the agent's training.
The reward defined in \eref{eq:reward} gives the agent a clear signal of whether it is getting closer to the target image faster or slower.
We compared both $l_2$ and Wasserstein distances to estimate $L_t$ in \eref{eq:reward} in the hope of accelerating learning with better reward shaping \cite{pmlr-v70-arjovsky17a}.
$l_2$ distance is used as a reference metric similar to the pixel-based metrics commonly used in computer vision.
Wasserstein distance is a probability distribution-based metric.
It is an estimate of the distance between two probability distributions.
The Wasserstein distance is estimated using the Sinkhorn iteration algorithm from a maximum entropy perspective between the two probability distributions \cite{NIPS2013_af21d0c9}.

Optimizing for pixel-error such as $l_2$ distance encourages finding pixel-wise averages for a plausible solution. 
It typically results in the loss of high-frequency details, giving overly smooth images with poor perceptual quality \cite{bruna2015super,ledig2017photo}.
Wasserstein distance allows one to better preserve the probability distribution of the light in the image \cite{pmlr-v70-arjovsky17a}.
Performances of the two metrics in training a DRL agent in the environment are presented in \sref{ssec:metric}.

\subsubsection{Percept quality in different virtual patients}
\label{sssec:experiment_different_patient}

\paragraph{Pixel-based versus perceptual-based comparison}

Agent-generated percepts are compared to the NSA algorithm described in \sref{ssec:nsa_stimulation_algorithm}.
The images are resized according to the Argus II layout of 10 by 6 electrodes.
The intensity of the pixels is uniformly assigned to the amplitude of the electrical pulse described in \sref{sssec:environment}.

We use the $l_2$ norm and the mean squared error (MSE), which are two \textit{de facto} standard in image restoration \cite{kautz2015loss} to compare percepts generated by the agent and the NSA algorithm.
We define the metrics for $M$ by $N$ images as follows:

\begin{equation}
    l_2 = \sum_{i=0,j=0}^{M,N}\sqrt{(I_{i,j} - P_{i,j})^{2}}
    \label{eq:l2_distance}
\end{equation}

\begin{equation}
    MSE = \sum_{i=0,j=0}^{M,N}\frac{(I_{i,j} - P_{i,j})^{2}}{MN}
    \label{eq:mse_distance}
\end{equation}

However, they do not correlate well with image quality as perceived by the human visual system \cite{kautz2015loss}.
Therefore, we use the mean structural similarity index measure (MSSIM), which is a perceptually motivated metric \cite{1284395}.
We calculate the MSSIM with non-negative image patches $\mathbf{x}$ and $\mathbf{y}$ of size 7 by 7 as follows :

\begin{equation}
    MSSIM(\mathbf{x},\mathbf{y}) = \frac{(2\mu_{\mathbf{x}}\mu_{\mathbf{y}} + c_{1})(2\sigma_{xy} + c_{2})}{(\mu_{\mathbf{x}}^{2}+\mu_{\mathbf{y}}^{2} + c_{1})(\sigma_{\mathbf{x}}^{2} + \sigma_{\mathbf{y}}^{2} + c_{2})}
\end{equation}
$c_1=0.01$ and $c_2=0.03$ are small constants that add numerical stability when the means $\mu$ or the standard deviation $\sigma$ are close to zero \cite{1284395}.
MSSIM takes into account the local characteristics of the image in a way similar to that of the human visual system.
On the contrary, pixel-based metrics, such as the $l_2$ norm and MSE, evaluate the difference between the corresponding pixels of two images independently of the nearby pixels.

\paragraph{Adaptation to different virtual patients}

Two SAC agents are trained on two virtual patients with different $\rho$ constant and $N$ single-electrode stimulation.
The spatial decay constant $\rho$ is varied to simulate two realistic virtual patients with different distances between the electrodes and the retina.
To ensure that a phosphene is anisotropic, its shape must be dominated by axon fascicle stimulation ($\lambda > \rho$).
The number of steps in an episode equivalent to the number of single-electrode stimulation $N$ also increases.
The high $\rho$ and high $N$ make the task more difficult for both approaches because the algorithms must deal with many large phosphenes.

\section{Results}
\label{sec:results}

\subsection{Effect of reward shaping on agent’s learning}
\label{ssec:metric}

\begin{figure}[t]
    \centering
    \includegraphics[width=0.5\textwidth]{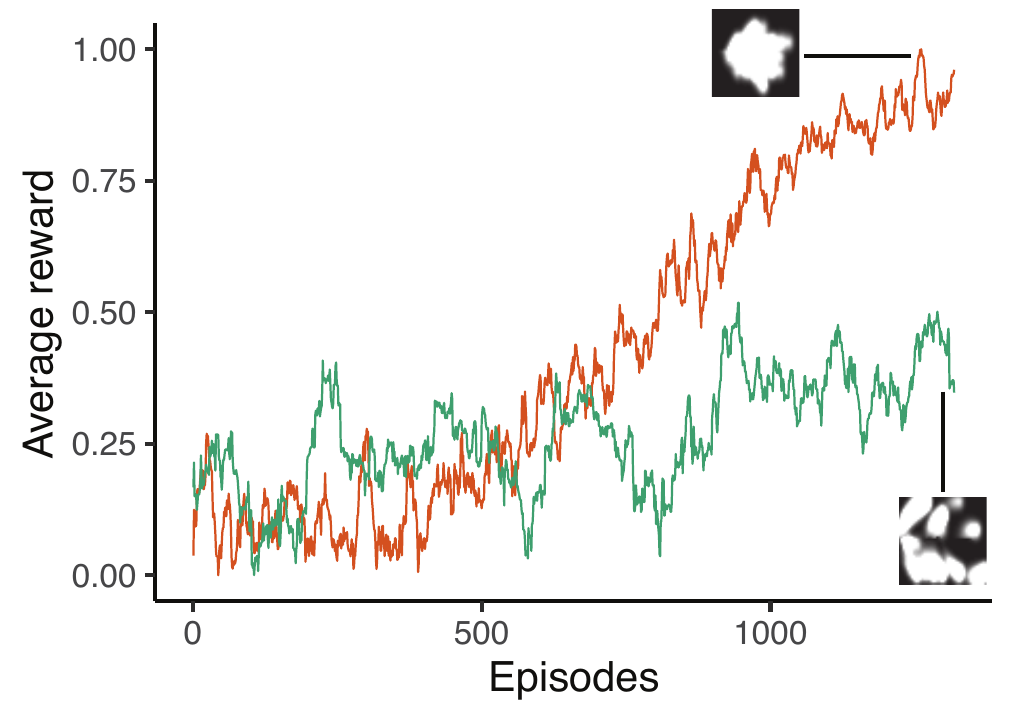}
    \caption{Comparison of the episode average reward of two SAC agents during training using respectively $l_2$ (red) and Wasserstein (green) based reward obtained by the SAC agent. Representative samples of percept produced by the agents early during training are shown. The reward average are normalized over a training iteration for the comparison.}
    \label{fig:euclidean_wasserstein_reward}
\end{figure}

\Fref{fig:euclidean_wasserstein_reward} shows that the reward estimated by the $l_2$ distance quickly saturates to a reward value after only 1000 episodes, while the reward estimated by the Wasserstein distance slowly progresses.
The results with a conventional $l_2$ distance as an estimator of $L_{t}$ are very limited.
Looking at the samples as shown in \fref{fig:euclidean_wasserstein_reward} reveals that the $l_2$ distance only taught the agent to represent low spatial frequencies of the dataset, while the Wasserstein distance preserves the particularity of each character.
The agent's reward then saturates as shown in \fref{fig:euclidean_wasserstein_reward} since it fails to learn to use the finer structures of the image to increase its reward.
The behavior persists in experiments (not presented here) that include data augmentation in an effort to remove bias from the MNIST dataset toward high-luminance in the center.

A Wasserstein-based reward stabilizes learning and allows more exploration without catastrophic failures as opposed to the $l_2$-based reward.
Wasserstein-based reward gives useful positional information regarding the distance of the phosphene from the high-luminance region of the image.
$l_2$ distance fails to give this information through the reward since it is calculated pixel-wise.
Therefore, the use of a Wasserstein-based reward results in more persistent electrode stimulation outside the high-luminance region.
Random single-electrode stimulation is added to \tref{tab:nsa_results} to better understand the results, as no other reference can be used as a benchmark.
Random stimulation helps to better grasp the range of values specific to each metric in the environment.
It also helps to contextualize pixel-based metrics with a tighter range, such as the $l_2$ norm and MSE, with perceptual metrics such as MSSIM.

\subsection{Percept quality in different virtual patients}
\label{ssec:percept_quality_in_different_virtual_patients}

The experiment presented in the previous section establishes that the best metric to evaluate the similarity between $I$ and $P$ is the Wasserstein distance as an estimator $L_{t}$.
Therefore, the percepts generated by the SAC agent in \fref{fig:mnist_sample} are obtained by training the agent with the Wasserstein distance.
In \sref{sssec:adaptation_different_virtual_patient}, the agent is trained in different virtual patients to demonstrate the flexibility of the proposed approach over the NSA stimulation algorithm.

\subsubsection{Pixel-based versus perceptual-based comparison}
\label{sssec:error_perceptual_error_comparison}

\Tref{tab:nsa_results} shows the image reconstruction metrics for the $\rho = 200$ and $\lambda = 500$ experiment in which the implant is close enough to the retina to obtain intelligible percepts with the NSA algorithm.
Samples of the percept generated for this virtual patient are shown in \fref{fig:mnist_sample_d01f9_low_rho}.
NSA single-electrode stimulations are, by default, limited to the high-luminance region of the image.
However, the SAC agent has the freedom to choose any single-electrode stimulation.
This is particularly visible when comparing the NSA and SAC agent samples in the second rows of \ref{fig:mnist_sample_d01f9_low_rho} and \ref{fig:mnist_sample_3e3eb_high_rho}.
This results in single-electrode stimulation in the out-of-high luminance region, thus increasing the $l_2$ distance and MSE for the SAC agent (See \tref{tab:nsa_results}).
However, the SAC agent significantly outperforms the NSA algorithm in preserving the structural integrity of the image measured with MSSIM ($p<0.0001$) as observed in \fref{fig:mnist_sample}.
NSA has significantly lower error-based metrics than the SAC agent ($l_2$; $p<0.0001$ and MSE: $p<0.0001$).
It shows that an SAC agent, despite higher values in error-based metrics, increases the readability of the digits, in contrast to the NSA algorithm.

\Table{\label{tab:nsa_results}Metric evaluating the percept and target image. Only the experiment in which NSA obtains readable percept ($\rho = 200$ and $\lambda = 500$) is shown. Lower $l_2$ and MSE is better. A higher MSSIM is better. 
The mean and standard deviation are calculated on 1000 MNIST images.}
         \br
         Metric & Random & NSA \cite{luo2016argus} & SAC Agent\\
         \mr
         $l_2$ norm & 13.54(1.67) & \textbf{11.69(1.63)} & 12.58(2.07) \\
         MSE & 0.11(0.02) & \textbf{0.08(0.02)} & 0.09(0.03) \\
         MSSIM & 0.07(0.05) & 0.28(0.08) & \textbf{0.35(0.11)} \\
         \br
\endTable   

\subsubsection{Adaptation to different virtual patients}
\label{sssec:adaptation_different_virtual_patient}

\begin{figure}[t]
    \begin{subfigure}[t]{0.5\textwidth}
        \includegraphics[width=\textwidth]{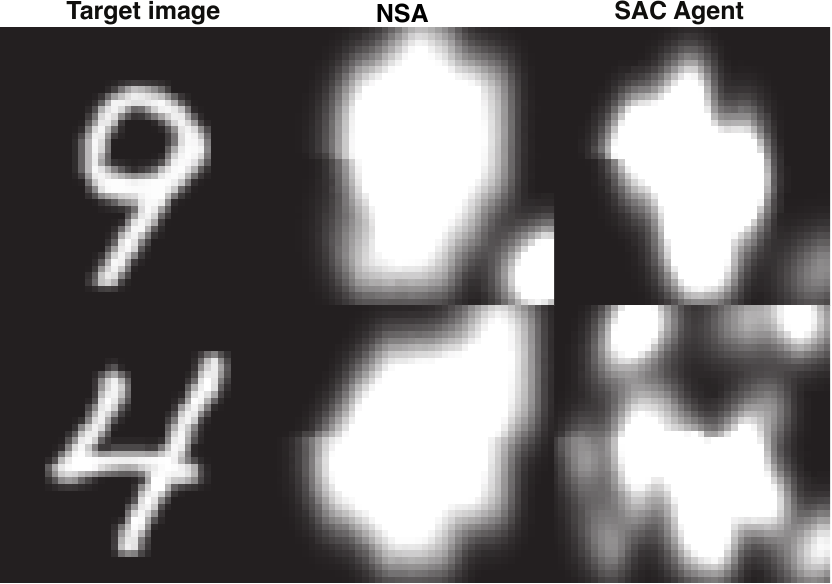} 
        \caption{}
        \label{fig:mnist_sample_d01f9_low_rho}
    \end{subfigure}
    \hfill
    \begin{subfigure}[b]{0.5\textwidth}
        \includegraphics[width=\textwidth]{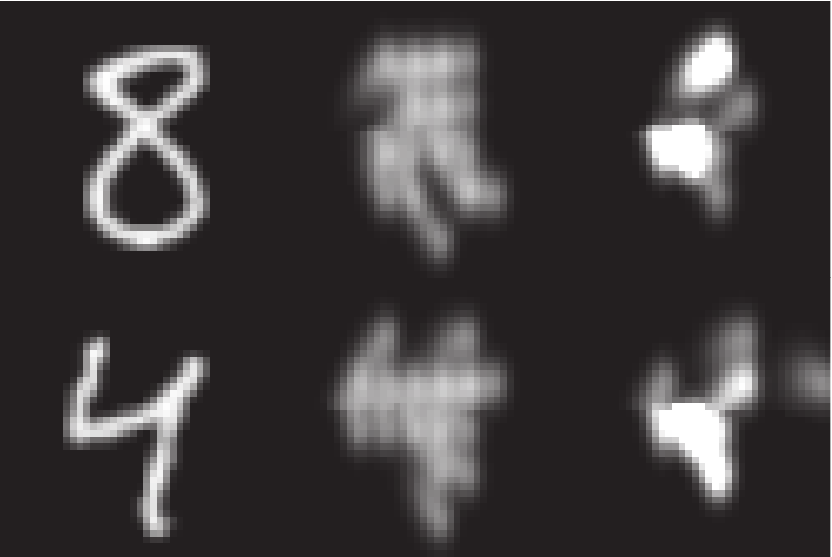} 
        \caption{}
        \label{fig:mnist_sample_3e3eb_high_rho}
    \end{subfigure}
    \caption{Two virtual patient's percepts with different axon map model and  single-electrode stimulation parameters. Columns from left to right are respectively the initial MNIST target image, the NSA generated percept and the SAC agent generated percept. (a) Samples generated with $\rho=315$, $\lambda=500$ and $N=32$. (b) Samples generated with $\rho=200$, $\lambda=500$ and $N=16$. Axon map model parameters are influenced by experiments realised with implanted subjects. Single-electrode stimulation is a biphasic pulse with a fixed amplitude of \SI{10}{\micro\ampere}.}
    \label{fig:mnist_sample}
\end{figure}

These experiments were carried out to demonstrate the limits of NSA algorithms in the condition often observed in implanted patients where the implant is relatively far from the retina \cite{beyeler2019model}.
A high distance between the stimulating electrode and the retina (or a high $\rho$) produces larger phosphenes that result in low resolution \cite{palanker2004migration}, as shown in \fref{fig:mnist_sample_3e3eb_high_rho}.
The NSA generated percepts are only readable with the implant close to the retina (low $\rho$) and with fewer single-electrode stimulations (low $N$), as observed by comparing \fref{fig:mnist_sample_d01f9_low_rho} and \fref{fig:mnist_sample_3e3eb_high_rho}.
SAC agent better preserves the readability of characters in \fref{fig:mnist_sample_3e3eb_high_rho} when the implant is further from the retina (high $\rho$) and a large number of single-electrode stimulations (high $N$). 
Therefore, the SAC agent can better adapt to more restrictive implant placements in patients in terms of resolution. 

\section{Discussion}
In this paper, we propose to address the anisotropic phosphene problem in epiretinal stimulation as an SBR task.
We present a DRL agent that can improve visual perception of numbers in the context of epiretinal stimulation.

We emphasize the fact that the reward design of the new environment and the evaluation of the agent's performances must take into account the perceptual nature of the task.
We address this limitation found in the reward design by comparing the $l_2$ and Wasserstein distances to assess the differences between the percept and the target image. 
The agent rewarded with the $l_2$ distance as an estimator uses only the electrodes near the center of the image, resulting in indistinguishable characters.
A Wasserstein-based reward thus outperforms the $l_2$-based reward, as it conveys information about the distance between the distribution of light of $I$ and $P$ rather than the pixel-sharp $l_2$-based reward.
This phenomenon is similar to the mode collapse phenomenon observed in GAN \cite{metz2016unrolled}.
Using a perceptual-based metric as a reward, such as Wasserstein, allows learning to stabilize and avoid mode collapse observed with a pixel-based metric.
More research is needed to assess whether the Wasserstein-based distance eliminates this phenomenon, as does the Wasserstein GAN algorithm \cite{wgan}.

The limitation of pixel-based metrics to evaluate the agent's performance is solved by proposing the more perceptually relevant MSSIM index.
These results align with other recent SBR approaches that account for the human perception instead of pixel-wise error metrics in the loss function \cite{bruna2015super,ledig2017photo,pmlr-v70-arjovsky17a}.
This is a first attempt to introduce human perception metrics in the design of a DRL stimulation algorithm for epiretinal implants.
Future work aimed at improving and restoring vision in patients should use metrics that account for human perception instead of methods that use pixel-based accuracy.

The SAC agent outperforms the NSA algorithm in preserving the similarity between target images and the generated percepts in different virtual patients.
It shows that the SAC agent adapts to different difficulty levels of the task as opposed to the NSA algorithm.
This is important in the context of the real patient.
It has been shown to be difficult to simultaneously modulate the size and brightness of phosphene \cite{nanduri2012frequency}.
The SAC agent allows one to circumvent certain limitations of the NSA algorithm, such as decreasing the intensity or inactivating electrodes in the case of anisotropic phosphenes or too large phosphenes.
Moreover, it automatically learns an optimal single-electrode stimulation sequence without human percept inspection and tuning.
It only requires standard hyperparameter tuning of the DRL agent.
However, previous conclusions are limited by the fact that single-electrode stimulations are assumed to be independent of each other based on the protocol used to develop the axon map model \cite{beyeler2019model}.

Limitations regarding the speed of the implementation of the axon map model during these experiments considerably slowed data generation.
Recent work \cite{10.1145/3519391.3524034} by the authors of the axon map model offers hope to replace the current implementation with a neural network.
This approach uses an approximation of the axon map model with a neural network which allows experiments to use only a GPU, eliminating the need for data transfer between the CPU and GPU memory.
This could significantly increase the percept generation process and the calculation of the reward in the proposed environment.
Therefore, accelerating the agent training process.

The training procedure could be further improved with imitation learning.
The agent currently learns from its interaction with the environment.
Using a dataset generated with the NSA algorithm could serve as a baseline training before training directly in the environment.
This could potentially improve the out-of-high luminance stimulation observed in the samples presented in \fref{fig:mnist_sample}.

\section{Conclusion}

In summary, this paper demonstrates that the formalization of epiretinal stimulation as an SBR problem allows for the full diversity of anisotropic phosphenes to be exploited, as opposed to the current NSA approach.
Previously unwanted phosphene shapes can now expand the complexity of possible percepts for patients with an epiretinal implant.
Further studies introducing metrics based on human perception into algorithm design could enhance the quality of recovered vision.
This allows us to better personalize the algorithm for each patient and create a better user experience.
This opens new ways to significantly improve the visual acuity of patients implanted with an epiretinal implant.

\ack

The support of the Digital Research Alliance of Canada and NSERC are gratefully acknowledged.

\section*{Declaration of competing interest}
The authors declare that they have no competing interests.

\section*{Data availability}
The MNIST \cite{deng2012mnist} dataset used in these experiments is obtained through the PyTorch Python library \cite{NEURIPS2019_9015}.
The proposed environment to reproduce the experiments is available at \href{https://github.com/NECOTIS/rlretina.git}{rlretina}.

\section*{References}

\bibliographystyle{vancouver}
\bibliography{reference.bib}

\end{document}